\documentclass{article}

    \PassOptionsToPackage{numbers, compress}{natbib}
 \usepackage[preprint]{neurips_2026}

\usepackage[utf8]{inputenc} 
\usepackage[T1]{fontenc}    
\usepackage{hyperref}       
\usepackage{url}            
\usepackage{booktabs}       
\usepackage{amsfonts}       
\usepackage{nicefrac}       
\usepackage{microtype}      
\usepackage{xcolor}         
\usepackage{graphicx}
\usepackage{booktabs}
\usepackage{amsmath}
\usepackage{amssymb}
\usepackage{multirow}
\usepackage[table]{xcolor}
\usepackage{adjustbox}
\usepackage{subcaption}
\usepackage{duckuments}
\usepackage{wrapfig}
\usepackage{booktabs}
\usepackage{pifont} 
\DeclareMathOperator*{\argmin}{arg\,min}

\newcommand{\psp}{\kern0.2ex}
\newcommand{\nsp}{\kern-0.1ex}

\title{\texorpdfstring{\!\!\!ComPose:\,When\:to\,Trust\:Hands\:for\,Object\:Pose\,Tracking}{ComPose: When to Trust Hands for Object Pose Tracking}}

\author{
  Jisu Shin\textsuperscript{1}, Junoh Lee\textsuperscript{1}, JunGyu Lee\textsuperscript{2}, Inhwan Bae\textsuperscript{3}, Dohyeon Lee\textsuperscript{2}, Hokyun Im\textsuperscript{2}, \\
  \textbf{Youngwoon Lee\textsuperscript{2}, ~Hae-Gon Jeon\textsuperscript{2}\thanks{Corresponding Author}}\\
  \textsuperscript{1}GIST~
  \textsuperscript{2}Yonsei Univ.~
  \textsuperscript{3}DGIST \\ 
  \texttt{\{jsshin98, juno\}@gm.gist.ac.kr}, \\
  \texttt{\{jungyu.lee, jellyho, youngwoon, earboll\}@yonsei.ac.kr}, \\
  \texttt{dohyeon@postech.ac.kr}, 
  \texttt{inhwanbae@dgist.ac.kr}
}

\begin{document}

\maketitle

\vspace{-5mm}

\begin{abstract}
Reconstructing the motion of objects from videos is a key component for embodied AI and robot manipulation. 
While diverse approaches to object pose tracking have been studied, they rely heavily on strong external priors, such as depth data or 3D templates, and remain highly vulnerable to severe occlusions by hand grasps despite the use of explicit masks.
In this work, we present ComPose, a 6DoF object tracking framework designed for hand-aware object pose estimation from RGB video. Rather than treating the hand purely as an occluder, our method harmonizes hand motions as a \textit{complementary cue} for object tracking. In detail, we recover a variety of object motions over time by combining object and hand cues from foundation models within a unified tracking pipeline.
Here, ComPose adaptively selects informative hand joints, combines object- and hand-derived cues for motion estimation, and refines the resulting object motion using visible geometric evidence and a learned correction.
We further enforce the temporal consistency over both rotation and translation, yielding stable 3D object trajectories over time without any external smoothing.
Extensive experiments show that our method is accurate, efficient, and robust under severe hand occlusion and geometric ambiguity.
In addition, the resulting trajectories can also effectively transfer to downstream robot manipulation by enabling robots to reconstruct human actions from online videos.

\end{abstract}

\vspace{-3mm}

\begin{figure}[htbp]
    \centering
    \includegraphics[width=\linewidth, page=1]{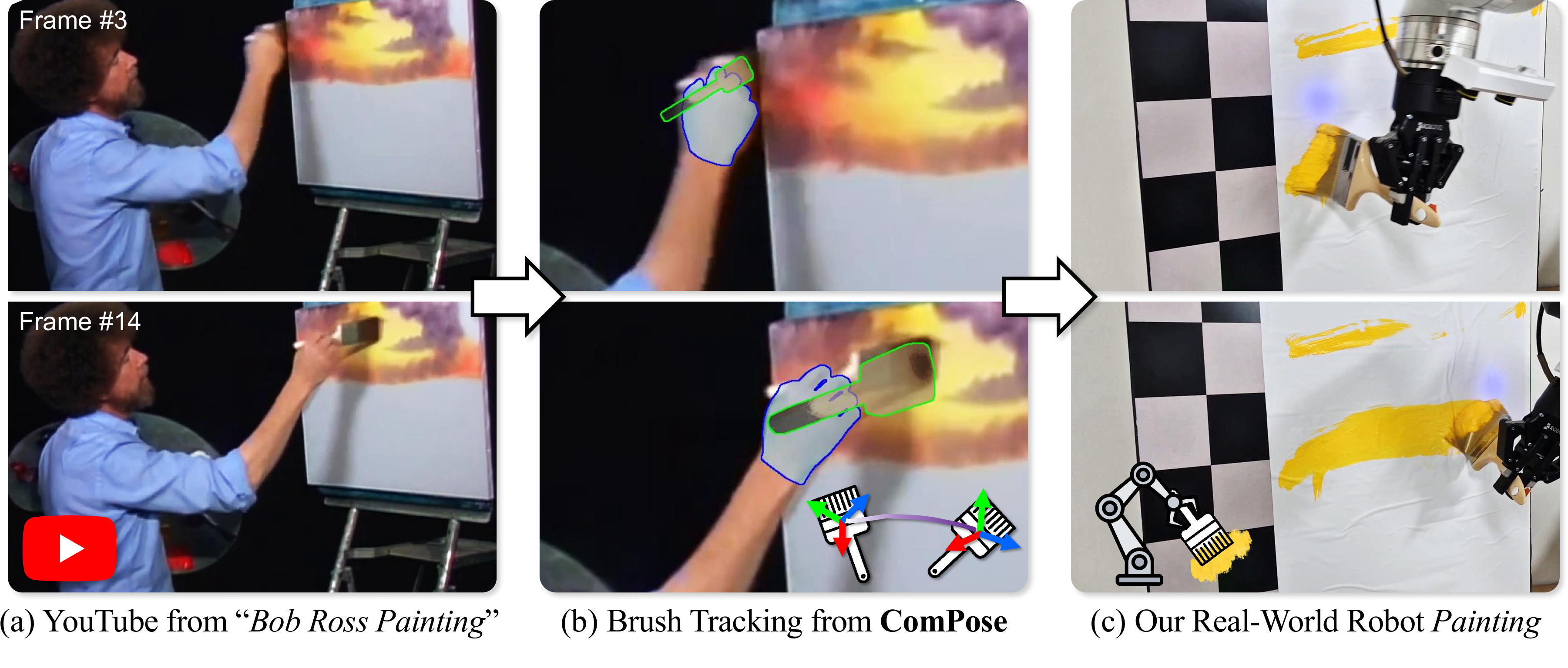}
    \vspace{-6mm}
    \caption{\textbf{Robot manipulation guided by the internet video.} 
    (a) Given an instructional painting video \textit{"Bob Ross Painting"} from Youtube, our method (b) predicts Bob's brush trajectory and (c) the estimated smooth trajectory is directly transferred to a real-world robot for painting.}
    \label{fig:teaser}
\end{figure}

\section{Introduction}
Object pose estimation and tracking from monocular RGB video is one of the core problems for understanding object motion and dynamic scene interactions in the real-world. Its importance has grown further with recent progress in embodied AI and robotics, where object motions observed in videos can serve as direct guidance for transferring human experts' demonstrations to robotic systems. 
More broadly, reliable tracking of manipulated objects enables structured understanding of scene dynamics and human-object interaction, making it an important component for video understanding, embodied perception and robot manipulation.

Existing object tracking and pose estimation methods often rely on strong external guidance, such as depth sensors~\cite{wen2021bundletrack,wen2023bundlesdf,wang2019densefusion}, object templates~\cite{li2018deepim,labbe2020cosypose}, or specialized correspondence modules~\cite{sun2022onepose,nguyen2024gigapose}. 
In particular, many 6DoF pose estimation methods assume given object-specific shape priors, such as CAD models or ground-truth meshes, which limits their generalizability in real-world settings.
Moreover, under hand-object interaction, matching- or correspondence-based methods often become unreliable because large portions of the object are occluded by the hand, leaving only limited visible regions for geometric reasoning.
As a result, recovering accurate object motion solely from RGB observations remains particularly challenging in hand-held object tracking scenarios, especially under hand-induced occlusion.


This implies that reliable hand-held object tracking should not depend on object observations alone.
When sufficient object regions are visible, object geometry provides direct evidence for tracking.
When the object is heavily occluded by hand, however, hand motion can offer complementary information about how the object moves.
The hand trajectory, orientation and grasp configuration physically constrain the motion of the manipulated object, making hand cues strong priors for motion estimation.
Motivated by this observation, we incorporate object geometry and hand pose within a unified tracking framework.

In this work, we propose \textbf{ComPose}, a \textbf{Com}plementary \textbf{Pose} tracking framework for hand-held object tracking that adaptively \textit{composes} object- and hand-derived cues. Our method is motivated by a simple question: \emph{when should object tracking trust the hand?} 
We leverage off-the-shelf 3D foundation models to recover dense 3D geometry without requiring explicit reconstruction pipelines or object-specific priors.
However, their outputs remain inherently partial, as they are derived only from visible regions and represented as point clouds rather than complete object geometry.
Consequently, geometry-based reasoning becomes unreliable when the object is heavily occluded by the hand.
At the same time, hand cues alone are also insufficient, as articulated hand motion does not always correspond to object motion, particularly before stable contact is established or during grasping.


To address this challenge, our adaptive tracking network predicts how object- and hand-derived cues should be used for pose estimation.
Specifically, it predicts a gating parameter that controls how much the final rotation should rely on the object cue or the hand cue, and a residual translation parameter that refines the initial translation estimate from visible object geometry.
A key challenge in using hand cues is that the articulated hand motion does not always reflect the motion of the manipulated object, moreover, the informativeness of each joint depends on the grasp configuration.
We therefore predict which hand joints are most informative for pose estimation, and use them to form a hand-based motion cue that complements the object-based estimate.
Notably, these joint weights are learned implicitly from the rotation supervision, without requiring any additional joint-level annotation or auxiliary loss.
For rotation, rather than predicting each frame independently, we estimate relative rotations between frames so that the prediction is grounded in frame-to-frame motion relationships.
For translation, the residual offset compensates for the bias introduced by partial visibility, since the initial estimate is computed only from the visible object region.
Finally, by imposing additional temporal consistency and smoothness constraints, our method produces stable and temporally coherent object trajectories over the full sequence.

Extensive experiments demonstrate that our approach yields accurate and temporally stable 3D object trajectories while remaining efficient in practice.
Our method improves robustness under severe hand occlusions and in geometrically ambiguous cases such as symmetric objects, where object-only tracking is often unstable~\cite{labbe2022megapose, ornek2024foundpose}.
Furthermore, the resulting trajectories transfer effectively to robotic manipulation without additional robot-specific trajectory optimization, showing that temporally consistent hand-held object tracking from videos can serve as a practical interface between human demonstrations and robot execution.

\noindent\textbf{Our contributions are summarized as follows:}
\vspace{-1mm}
\begin{itemize}
    \item We present a framework for hand-held object tracking that enables temporally consistent 3D object trajectory estimation directly from RGB videos, without any external guidance.
    \item  We leverage the hand as a complementary cue under occlusion and propose an adaptive module that selects informative joints and fuses hand and object cues.
    \item We achieve strong tracking performance and show effective transfer of the predicted smooth trajectories to downstream robot manipulation.
\end{itemize}
\vspace{-3mm}

\section{Related Work}\label{sec:related_work}

\subsection{6D Object Pose Estimation via Semantic Priors}
Pioneering works directly estimate the position and orientation of objects in an RGB image through end-to-end regression~\cite{kendall2015posenet,xiang2017posecnn,do2018deep,chen2022epro}, 2D detection integrated with dense matching~\cite{zakharov2019dpod,shugurov2021dpodv2} or template matching~\cite{kehl2017ssd,shugurov2022osop, moon2025co}. To further improve regression performance, various strategies apply iterative refinements through render-and-compare optimizations~\cite{li2018deepim, labbe2020cosypose,xu2022rnnpose,iwase2021repose,lipson2022coupled,labbe2022megapose, ornek2024foundpose}, occlusion handling through segmentation and self-occlusion modeling~\cite{hu2019segmentation,di2021so}, or leverage synthetic training data and bridge the gap using self-supervised methods with pseudo-labels~\cite{hai2023pseudo} or with implicit 3D representation rendering~\cite{yen2021inerf, park2020latentfusion}. 

To overcome the inherent instability of direct regression, local feature mapping approaches predict semantic keypoints~\cite{pavlakos20176,peng2019pvnet,castro2023crt} or 3D bounding box corners~\cite{rad2017bb8, hai2023rigidity}. Techniques, which map pixel-wise dense coordinates or discretize object surfaces into fragments~\cite{hodan2020epos,wang2021gdr,su2022zebrapose,park2019pix2pose}, have demonstrated significant resilience to severe occlusions. Additionally, notable advancements in robust pose estimation have been made from coordinate-based disentangled networks~\cite{li2019cdpn} and multi-representation fusion~\cite{song2020hybridpose}. Moreover, performance has been further advanced by introducing foundation model prompts~\cite{fan2024pope}.
We note that approaches, which use additional depth information from RGB-D images to capture global geometric structures~\cite{wang2019densefusion, xu2018pointfusion, he2020pvn3d,he2021ffb6d,li2023depth,jiang2022uni6d, chen2020g2l,wang20206,huang2025cap,liu2022catre}, are out-of-scope to our work.

\subsection{Generalization to Novel Objects and Video Tracking} 

Since accurate 3D object models are not always available, a significant line of research focuses on generalizing pose estimation to novel objects without relying on predefined template models like CAD. Pioneering works achieve this by generalizing across intra-class variations using canonical object representations~\cite{wang2019normalized}. Subsequent studies have further enabled models to accommodate complex structural deviations within specific categories~\cite{tian2020shape,chen2021sgpa,weng2021captra,lin2022category,lin2021dualposenet,irshad2022centersnap,li2023generative,lin2022single,zhang2022rbp,chen2020category,zhang2023genpose,nguyen2024gigapose}.

Recently, the paradigm has shifted toward fully model-free configurations. To break beyond intra-class boundaries, frameworks directly reconstruct 3D feature correlations from sparse reference images or short video scans~\cite{sun2022onepose,he2022onepose++,liu2022gen6d}. These 3D model-free approaches have naturally extended into the temporal domain for novel-object 6D pose tracking~\cite{wen2021bundletrack,wen2023bundlesdf}. Furthermore, state-of-the-art methods leverage temporal consistency to jointly optimize object meshes and 6DoF trajectories from RGB video~\cite{song2025prior}, while others achieve fast novel object estimation via single-view retrieval~\cite{nguyen2024gigapose}. Most recently, the integration of foundation models~\cite{lin2024sam,liu2025one2any,lee2025any6d} has significantly advanced tracking stability in cluttered environments~\cite{wen2024foundationpose,ponimatkin20256d}.
However, RGB-only methods suffer fundamentally from the lack the stability required for downstream robotic manipulation. To address this, we propose a framework that achieves robust alignment by leveraging the hand poses, thereby producing accurate and temporally consistent results. 




\section{Methodology}
Given an input video of hand-object interaction, our goal is to estimate the motion of the manipulated object over time. The proposed framework consists of three components: a 3D foundation model-based tracking, a hand pose estimator-based tracking, and ComPose.
Based on dense 3D geometry from the foundation model (Sec.~\ref{sec:3DFoundation}) and 2D--3D hand joints from the hand estimator, we first obtain object- and hand-based pose cues using the proposed formulation (Sec.~\ref{subsec:estimation}), and then combine them according to their estimated reliability through ComPose (Sec.~\ref{subsec:blending}).

\begin{figure}[t]
    \centering
    \includegraphics[clip, width=\linewidth]{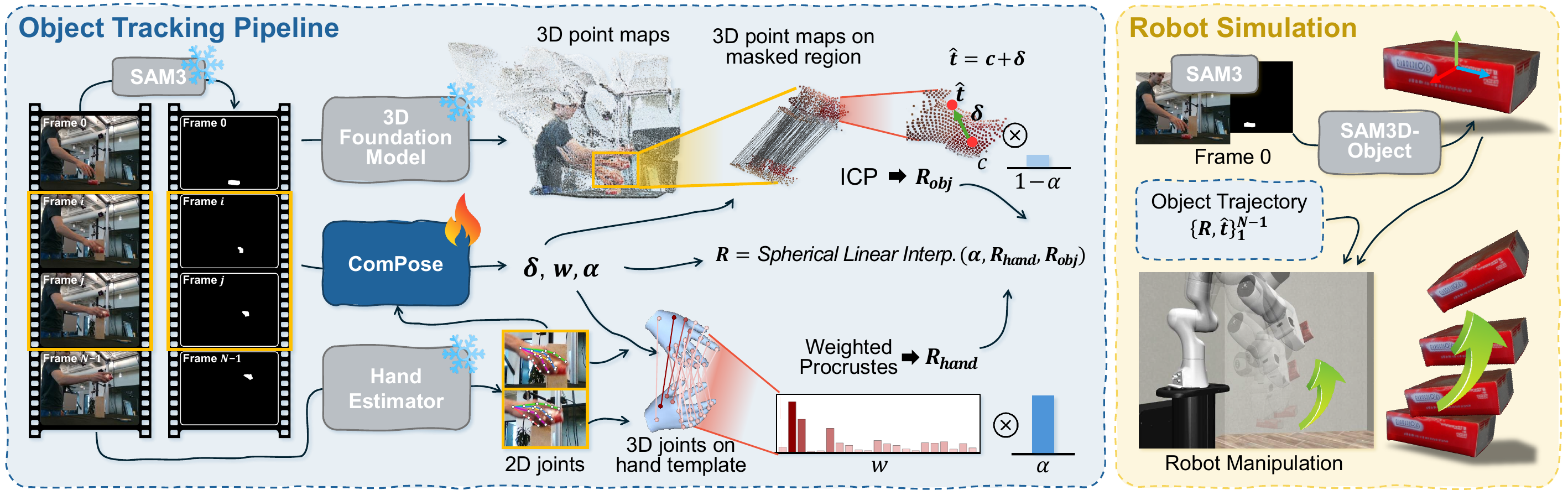}
    \vspace{-5mm}
    \caption{An illustration of the proposed object tracking pipeline and its application to robot simulation.}
    \label{fig:placeholder}
\end{figure}

\subsection{Preliminary: 3D Foundation Models}
\label{sec:3DFoundation}

Recent feed-forward visual geometry networks~\cite{wang2025vggt, elflein2026vgg, wang2025pi, cong2026flow3r, moreau2025off} enable dense 3D reconstruction from casual video.
Given $N$ input frames $\{I_i\}_{i=1}^N$, these models jointly estimate cross-frame consistent dense point maps, depth maps, and camera poses through a unified multi-view transformer $f_{\theta}$:
\begin{equation}
  \{P_i, D_i, C_i\}_{i=1}^N = f_{\theta}\left(\{I_i\}_{i=1}^N\right),
\end{equation}
where $P_i \in \mathbb{R}^{H \times W \times 3}$ is point map, $D_i \in \mathbb{R}^{H \times W}$ refers to depth, and $C_i \in SE(3)$ denotes the camera pose of the $i$-th frame. 
In this work, we build on these predictions as the primary geometric representation for object tracking. Based on these dense 3D outputs, our method estimates object pose through our pipeline (Sec.~\ref{subsec:estimation}).

\subsection{Pose Tracking}
\label{subsec:estimation}
Here, we combine two complementary tracking formulations: a template-free geometric formulation derived from dense 3D object observations, and a hand-based geometric formulation that uses hand motion as an auxiliary cue when object evidence becomes unreliable.
These complementary formulations are integrated into our adaptive tracking framework in Sec.~\ref{subsec:blending}.

\noindent \textbf{Object Template-Free Pose Tracking.}
Given two frames $I_i$ and $I_{j}$, we first obtain per-frame object masks using SAM3~\cite{carion2025sam3segmentconcepts}. 
We then extract dense 3D point maps from a pre-trained 3D foundation model and retain only the object-masked regions to obtain per-frame point clouds $P'_i \in \mathbb{R}^{K_i \times 3}$ and $P'_{j} \in \mathbb{R}^{K_{j} \times 3}$, where $K_i$ and $K_{j}$ denote the number of points reconstructed from the two masked frames.
Since explicit point correspondences are unavailable across frames, we estimate the relative object rotation $R_{\mathrm{obj}}$ using Iterative Closest Point (ICP)~\cite{besl1992method}:
\begin{equation}
  R_{\mathrm{obj}} = \argmin_{R \in \mathrm{SO}(3)} \sum_{k \in K_i} \left\| R(\mathbf{p}_k - \bar{\mathbf{p}}) - (\mathbf{q}_{\pi(k)} - \bar{\mathbf{q}}) \right\|^2,
\end{equation}
where $\mathbf{p}_k \in P'_i$ denotes a source 3D point, $\mathbf{q}_{\pi(k)} \in P'_{j}$ is its assigned closest point in the next frame, $\bar{\mathbf{p}}$ and $\bar{\mathbf{q}}$ are the centroids of $P'_i$ and $P'_{j}$, respectively, and $\pi(k)$ denotes the closest-point assignment updated iteratively during ICP.

In our implementation, we use Flow3R~\cite{cong2026flow3r} as the 3D foundation model. However, the proposed formulation does not rely on a specific model and can be applied to other 3D foundation models as well, which is validated additional comparisons in Sec.~\ref{subsec:exp}.

\noindent\textbf{Hand-based Pose Tracking.}
While the object transformation can be estimated from object point clouds alone, this estimation becomes less reliable when the object is heavily occluded by the hand. To address this issue, our idea is to exploit hand motion as an explicit cue. Since object surfaces do not provide explicit correspondences across frames, whereas hand joints do through the fixed skeletal topology of MANO~\cite{romero2022embodied}, we estimate object and hand transformations using different alignment strategies.
We recover 3D hand joint positions $\mathbf{J}_i, \mathbf{J}_{j} \in \mathbb{R}^{21 \times 3}$ using a hand pose estimator, where the $l$-th joint in frame $i$ is denoted by $\mathbf{J}_{l,i} \in \mathbb{R}^3$.
Because hand joints have known correspondences across frames, we estimate the relative hand rotation $R_{\mathrm{hand}}$ using weighted Procrustes alignment~\cite{gower1975generalized, lissitz1976solution, choy2020deep}:
\begin{equation}
\label{hand_proc}
  R_{\mathrm{hand}} = VU^\top, \quad
  \text{where} \quad
  U \Sigma V^\top =
  \mathrm{SVD}\!\left(
  \sum_{l=1}^{21} w_l \,
  (\mathbf{J}_{l,i} - \bar{\mathbf{J}}_i)
  (\mathbf{J}_{l,j} - \bar{\mathbf{J}}_{j})^\top
  \right),
\end{equation}
where $\bar{\mathbf{J}}_i$ and $\bar{\mathbf{J}}_{j}$ denote the anchor joint positions (the wrist joint is used in our implementation), and the weights $w$ are predicted by our complementary estimation module described in Sec.~\ref{subsec:blending}. These weights allow the model to identify which joints are most informative for object motion in a given frame pair, rather than treating all joints equally.
In particular, joints involved in object contact or stable grasp formation tend to provide stronger evidence for rigid object motion, whereas other joints may mainly reflect hand articulation that is less relevant to tracking.

In our implementation, we use WiLoR~\cite{potamias2025wilor} as the default hand pose estimator.
However, the proposed formulation is not tied to a specific hand model, which is also addressed in Sec.~\ref{subsec:exp}.

\subsection{ComPose: Complementary Pose Tracking using Object and Hand Cues}
\label{subsec:blending}

Neither object geometry nor hand motion alone provides a consistently reliable cue for rotation estimation in hand-object interaction, as already discussed in Sec.~\ref{subsec:estimation}.
Object-based estimation degrades when the visible object region is heavily occluded, whereas hand-based estimation can be inaccurate when articulation or re-grasping happens.
To address this, we introduce a complementary pose estimation module that jointly reasons over image observations and hand motion, and predicts both joint-wise confidence weights and a blending coefficient for adaptive hand-object fusion.

Given a temporal window of $N$ frames, we first extract 2D and 3D hand joints using WiLoR~\cite{potamias2025wilor}. 
In parallel, we obtain dense image features from DINOv2~\cite{oquab2023dinov2} and fuse them with the object mask, which are then processed by alternating local self-attention within each frame and global self-attention across frames~\cite{wang2025vggt}. 
Along with this, we extract joint tokens from the 2D hand joint keypoints and update them through cross-attention with the image tokens, followed by global self-attention across all frames. 
This design allows the model to identify which joints move consistently with the manipulated object and which joints are less reliable due to articulation or unstable contact. It also enables adaptive selection between hand and object cues for each frame pair.

For each temporal pair $(i,j)$ within the sampled window, the module predicts per-joint weights $w_{ij}$ and a blending coefficient $\alpha_{ij}\in[0,1]$. 
The predicted weights are used in the weighted Procrustes alignment (Eq.~\ref{hand_proc}) to obtain the hand-based relative rotation estimate \(R_{\mathrm{hand},\, ij}\). 
The blending coefficient determines how much the final rotation should rely on the hand cue.
We combine the object- and hand-based relative rotations by spherical linear interpolation in the unit-quaternion space:
\begin{align}
\hat{q}_{ij}
&=
\frac{\sin\big((1-\alpha_{ij})\theta_{ij}\big)}{\sin \theta_{ij}}\, q_{\mathrm{obj},\, ij}
+
\frac{\sin\big(\alpha_{ij}\theta_{ij}\big)}{\sin \theta_{ij}}\, q_{\mathrm{hand},\, ij}, \\
\theta_{ij}
&=
\arccos\!\left(
\left|
\left\langle q_{\mathrm{obj},\, ij}, q_{\mathrm{hand},\, ij}\right\rangle
\right|
\right).
\end{align}
and convert the interpolated quaternion back to the relative rotation \(\hat{R}_{ij}\). 
This adaptive fusion allows the method to rely more on object geometry when the masked object point cloud is reliable, and to shift toward hand motion when geometric evidence becomes ambiguous.

For translation, we use the centroid of the masked object point cloud as an initial estimate and predict only a residual correction per frame:
\begin{equation}
\hat{\mathbf{t}}_i = \mathbf{c}_i + \boldsymbol{\delta}_i,
\qquad
\mathbf{c}_i = \frac{1}{|P'_i|}\sum_{\mathbf{p}\in P'_i}\mathbf{p}.
\end{equation}
where $P'_i$ denotes the object-masked 3D point cloud and $\boldsymbol{\delta}_i$ is predicted from the pooled image feature together with the object centroid $\mathbf{c}_i$. This residual formulation stabilizes translation estimation while compensating for centroid bias caused by partial visibility and hand occlusion.



\noindent \textbf{Training Losses.}
We train on sampled three-frame windows $a<b<c$ from the input video, where the temporal gaps between sampled frames may vary. We supervise pairwise rotations over $\mathcal{S}=\{(a,b),(b,c),(a,c)\}$ and projected translations in image space, which is more stable under the scale ambiguity of monocular RGB observations, by $\mathcal{L}_{\mathrm{rot}}$ using rotation loss and $\mathcal{L}_{\mathrm{trans}}$ using L1 loss, respectively.

We further impose cumulative consistency $\mathcal{L}_{\mathrm{cons}}$ and temporal smoothness $\mathcal{L}_{\mathrm{smooth}}$:
\begin{equation}
\begin{aligned}
  \mathcal{L}_{\mathrm{cons}}
  &=
  \left\|
    \log\!\left(
      (\hat{R}_{bc}\hat{R}_{ab})^\top \hat{R}_{ac}
    \right)
  \right\|_2, \qquad
  \mathcal{L}_{\mathrm{smooth}}
  &=
  \left\|
  \frac{\hat{\mathbf{t}}_c-\hat{\mathbf{t}}_b}{\,c-b\,}
  -
  \frac{\hat{\mathbf{t}}_b-\hat{\mathbf{t}}_a}{\,b-a\,}
  \right\|_2.
\end{aligned}
\end{equation}
where $\hat{R}_{uv}$ denotes the relative rotation estimate for frame pair $(u,v)$.
The smoothness term penalizes changes in gap-normalized translation velocities under variable frame sampling.

To prevent the learned translation correction from drifting excessively away from the object point-cloud centroid, we impose a soft bound regularizer $\mathcal{L}_{\mathrm{bound}}$:
\begin{equation}
  \mathcal{L}_{\mathrm{bound}}
  =
  \left(
    \max\!\left(0,\|\boldsymbol{\delta}_i\| - 0.5\,\sigma_i\right)
  \right)^2,
\end{equation}
where $\boldsymbol{\delta}_i \in \mathbb{R}^3$ is the predicted translation residual for frame $i$, and $\sigma_i$ denotes the standard deviation of the object point cloud $P'_i$, used as a scale proxy.

The final objective function $\mathcal{L}$ is defined as follows:
\begin{equation}
  \mathcal{L}
  =
  \mathcal{L}_{\mathrm{rot}}
  + \lambda_t \mathcal{L}_{\mathrm{trans}}
  + \lambda_c \mathcal{L}_{\mathrm{cons}}
  + \lambda_s \mathcal{L}_{\mathrm{smooth}}
  + \lambda_b \mathcal{L}_{\mathrm{bound}},
\end{equation}
where $\lambda_t$, $\lambda_c$, $\lambda_s$, and $\lambda_b$ are empirically set to $10.0$, $1.0$, $1.0$, and $1.0$, respectively.
\section{Experiments}
\subsection{Experimental Settings}
\label{subsec:exp_setting}
\noindent \textbf{Datasets.}
We train on DexYCB~\cite{chao2021dexycb} S3 and HOI4D~\cite{liu2022hoi4d}, which together provide diverse view and egocentric hand-object interaction videos with diverse manipulation dynamics.
For evaluation, we use the DexYCB S3 test split for in-domain generalization, and HO-3D v2~\cite{hampali2020honnotate} and OakInk-v1~\cite{yang2022oakink} for cross-dataset evaluation.
For OakInk, we exclude bimanual handover and articulated-object sequences to match our single-hand tracking setting.

\noindent \textbf{Metrics.}
We evaluate object trajectories using both absolute pose accuracy and relative motion accuracy, following trajectory evaluation protocols commonly used in SLAM and visual odometry~\cite{sturm2012benchmark,zhang2018tutorial}.\footnote{We report ARE/RRE instead of ADD/AR-style metrics because our setting focuses on video-based object tracking rather than single-frame pose evaluation.
These metrics more directly capture trajectory accuracy and temporal consistency for downstream robot manipulation.} 
For relative motion accuracy, we report Relative Translation Error (RTE) and Relative Rotation Error (RRE). 
These relative metrics quantify how accurately the predicted trajectory captures motion between consecutive frames, thereby reflecting drift and temporal inconsistency.
For absolute pose accuracy, we report Absolute Trajectory Error (ATE) for translation and Absolute Rotation Error (ARE).
We report RRE and ARE in degrees, and RTE and ATE in the native scale of each dataset.

We additionally report the Temporal Correlation Coefficient (TCC), computed as the Pearson correlation between predicted and ground-truth per-frame motion changes.
We compute TCC separately for translation and rotation, and average the results across axes.
Let
\(
\log(\Delta R_i), \log(\Delta \hat{R}_i) \in \mathbb{R}^3
\)
denote the rotation-vector representations of the relative rotations.
We define
\begin{align}
  \mathrm{TCC}_\mathrm{R}
  &=
  \frac{1}{3}
  \sum_{d \in \{x,y,z\}}
  \rho \Bigl(
    \{(\log(\Delta \hat{R}_i))_d\}_{i=1}^{N-1},
    \{(\log(\Delta R_i))_d\}_{i=1}^{N-1}
  \Bigr), \\
  \mathrm{TCC}_\mathrm{T}
  &=
  \frac{1}{3}
  \sum_{d \in \{x,y,z\}}
  \rho \Bigl(
    \{(\Delta \hat{\mathbf{t}}_i)_d\}_{i=1}^{N-1},
    \{(\Delta \mathbf{t}_i)_d\}_{i=1}^{N-1}
  \Bigr),
\end{align}
where $\rho(\cdot,\cdot)$ denotes the Pearson correlation coefficient, and $d \in \{x,y,z\}$ specifies the three coordinate axes.
Complementary to error-based metrics, TCC measures whether the predicted trajectory follows the same temporal motion pattern as the ground truth. Higher values indicate better temporal alignment: $1$ denotes perfect positive correlation, $0$ indicates no correlation, and $-1$ denotes the perfect negative correlation with its ground truth.


 
\newcommand{\TCCR}{TCC\textsubscript{R}\!}
\newcommand{\TCCT}{TCC\textsubscript{T}\!}

\begin{table}[t]
    \centering
    \caption{Main Quantitative Comparison. Time shows the processing time per frame ((Pre-processing) Model Inference Time). \textsuperscript{$\dagger$ }: CAD model used.}
    \label{tab:main}
    \setlength{\tabcolsep}{0.5pt}
    \begin{adjustbox}{width=\columnwidth}
    \begin{tabular}{l|ccccccc|ccccccc|ccccccc}
    \toprule
        \textit{Dataset} & \multicolumn{7}{c|}{DexYCB} & \multicolumn{7}{c|}{HO-3D V2} & \multicolumn{7}{c}{OakInk-V1} \\  
        \midrule
        Method & RRE$\downarrow$ & RTE$\downarrow$ & ARE$\downarrow$ & ATE$\downarrow$ & \TCCR$\uparrow$ & \TCCT$\uparrow$ & Time & RRE$\downarrow$ & RTE$\downarrow$ & ARE$\downarrow$  & ATE$\downarrow$ & \TCCR$\uparrow$ & \TCCT$\uparrow$ & Time & RRE$\downarrow$ & RTE$\downarrow$ & ARE$\downarrow$ & ATE$\downarrow$ & \TCCR$\uparrow$ & \TCCT$\uparrow$ & Time \\ 
        \midrule
        MegaPose\textsuperscript{$\dagger$}\cite{labbe2022megapose}  & 46.1 & 2.1 & \cellcolor{yellow!30}{62.6} & \cellcolor{yellow!30}{5.0} & \cellcolor{yellow!30}{0.03} & \cellcolor{yellow!30}{0.270} & \cellcolor{yellow!30}{(0 / 610)} &  \cellcolor{yellow!30}{16.9} & \cellcolor{yellow!30}{0.5} & \cellcolor{red!30}{42.6} & \cellcolor{red!30}{2.4} & 0.02 & \cellcolor{yellow!30}{0.22} & (0 / 34K) & \cellcolor{yellow!30}{64.6} & \cellcolor{yellow!30}{1.8} & 88.2 & \cellcolor{orange!30}{4.3} & \cellcolor{yellow!30}{0.01} & \cellcolor{yellow!30}{0.19} & \cellcolor{orange!30}{(0 / 171)} \\
        FoundPose\textsuperscript{$\dagger$}\cite{ornek2024foundpose} & 55.0 & 4.9 & 66.5 & 6.2 & 0.02 & 0.14 & (38K / 5) & 33.4 & 1.8 & \cellcolor{orange!30}{46.0} & \cellcolor{orange!30}{2.7} & 0.02 & 0.10 & (39K / 239) & 74.7 & 3.2 & 92.8 & 5.6 & 0.01 & 0.06 & (75K / 32)\\ \midrule
        FreePose~\cite{ponimatkin20256d} & \cellcolor{orange!30}{2.0} & \cellcolor{orange!30}{0.7} & \cellcolor{orange!30}{31.2} & \cellcolor{orange!30}{4.2} & \cellcolor{orange!30}{0.20} & \cellcolor{orange!30}{0.51} & (0 / 752) & \cellcolor{orange!30}{2.6} & \cellcolor{red!30}{0.3} & 80.3 & \cellcolor{yellow!30}{3.4} & \cellcolor{orange!30}{0.15} & \cellcolor{orange!30}{0.24} & \cellcolor{yellow!30}{(0 / 5K)} & \cellcolor{orange!30}{2.7} & \cellcolor{red!30}{0.8} & \cellcolor{orange!30}{30.4} & \cellcolor{yellow!30}{5.1} & \cellcolor{orange!30}{0.15} & \cellcolor{orange!30}{0.27} & \cellcolor{yellow!30}{(0 / 1K)}\\
        UniHope~\cite{wang2025unihope} & \cellcolor{yellow!30}{44.2} & \cellcolor{yellow!30}{2.1} & 100.4 & 12.3 & 0.01 & 0.02 & \cellcolor{red!30}{(0 / 6)} & 31.8 & \cellcolor{orange!30}{0.4} & 76.9 & 4.3 & \cellcolor{yellow!30}{0.07} & 0.04 & \cellcolor{red!30}{(0 / 0.2)} & - & - & - & - & - & - & -\\
        Ours & \cellcolor{red!30}{1.5} & \cellcolor{red!30}{0.6} & \cellcolor{red!30}{19.5} & \cellcolor{red!30}{3.0} & \cellcolor{red!30}{0.55} & \cellcolor{red!30}{0.57} & \cellcolor{orange!30}{(33 / 0.9)} & \cellcolor{red!30}{2.4} & \cellcolor{red!30}{0.3} & \cellcolor{yellow!30}{49.2} & \cellcolor{orange!30}{2.7} & \cellcolor{red!30}{0.41} & \cellcolor{red!30}{0.40} & \cellcolor{orange!30}{(142 / 9)} & \cellcolor{red!30}{2.4} & \cellcolor{orange!30}{1.2} & \cellcolor{red!30}{28.8} & \cellcolor{red!30}{3.2} & \cellcolor{red!30}{0.19} & \cellcolor{red!30}{0.32} & \cellcolor{red!30}{(27 / 0.8)}\\
    \bottomrule
    \end{tabular}
    \end{adjustbox}
\end{table}

\noindent \textbf{Implementation Details.}
We train our model on an NVIDIA RTX PRO 6000 Blackwell GPU with a batch size of $128$ using the AdamW~\cite{loshchilov2017decoupled} with an initial learning rate of $1\times10^{-4}$ and weight decay of $1\times10^{-4}$. The learning rate is decayed following a cosine annealing schedule over $30$ epochs. 
At inference time, we obtain the object mask using SAM3~\cite{carion2025sam3segmentconcepts}, and estimate the initial object pose and mesh from the first frame using SAM3D-Object~\cite{chen2025sam}.
The estimated pose is used to initialize object tracking, while the reconstructed mesh is used for robot simulation. To reduce the cumulative rotation drift arising from sequential relative pose prediction, we additionally perform periodic re-anchoring every $K$ frames using wider-window anchor estimates, and distribute the resulting correction through spherical linear interpolation to avoid abrupt discontinuities.

For fair evaluation, we align all predictions, including both ours and the baseline methods, to the ground truth before computing pose errors.
For rotation, we align the first frame to remove coordinate-frame ambiguity.
For translation, we apply Umeyama alignment~\cite{umeyama2002least} over the full trajectory to account for global scale and reference-frame mismatch.
More details on datasets, metrics, and implementation details are provided in the supplementary material.

\begin{wraptable}{r}{0.43\columnwidth}
\vspace{-12mm}
\caption{Input requirements across models.\,CoT:\,Co-Tracker;\,V:\,video; $^{*}$\,reference-image-based option; $^{\ddagger}$\,retrieved CAD.}
\vspace{-2mm}
\label{tab:assumption_compare}
\centering
\setlength{\tabcolsep}{2.5pt}
\resizebox{\linewidth}{!}{
\begin{tabular}{lcccc}
\toprule
Method\hspace{6em} & RGB-D & CAD & ~CoT~ & Frames \\
\midrule
BundleTrack\cite{wen2021bundletrack}       & \checkmark &  &  & V \\
FoundationPose\cite{wen2024foundationpose} & \checkmark & \checkmark$^{*}$ &  & 1/V \\
\midrule
MegaPose\cite{labbe2022megapose}           &  & \checkmark &  & 1 \\
FoundPose\cite{ornek2024foundpose}         &  & \checkmark &  & 1 \\
FreePose\cite{ponimatkin20256d}            &  & \checkmark$^{\ddagger}$ & \checkmark & V \\
UniHOPE\cite{wang2025unihope}              &  &  &  & 1 \\
\textbf{Ours}  &  &  &  & V \\
\bottomrule
\end{tabular}
}
\vspace{-5mm}
\end{wraptable}

\subsection{Quantitative and Qualitative Comparisons}
\label{subsec:exp}
We compare our method with both model-based and model-free object pose estimation approaches.
As summarized in Tab.~\ref{tab:assumption_compare}, prior methods differ in their assumptions on RGB-D input, CAD models, external tracking, and the number of input frames.
MegaPose~\cite{labbe2022megapose} and FoundPose~\cite{ornek2024foundpose} require object CAD models, while UniHOPE~\cite{wang2025unihope} does not.
FreePose~\cite{ponimatkin20256d} instead retrieves a similar CAD model from a large-scale reference database~\cite{deitke2023objaverse,downs2022google} and refines trajectories using CoTracker~\cite{karaev2024cotracker} and PnP~\cite{lepetit2009ep}.
In contrast, our method requires neither CAD models nor retrieved shape priors or reference images of the target object.
Depth priors are beyond the scope of this work.

\begin{figure}[t]
    \centering
    \includegraphics[clip, width=\linewidth]{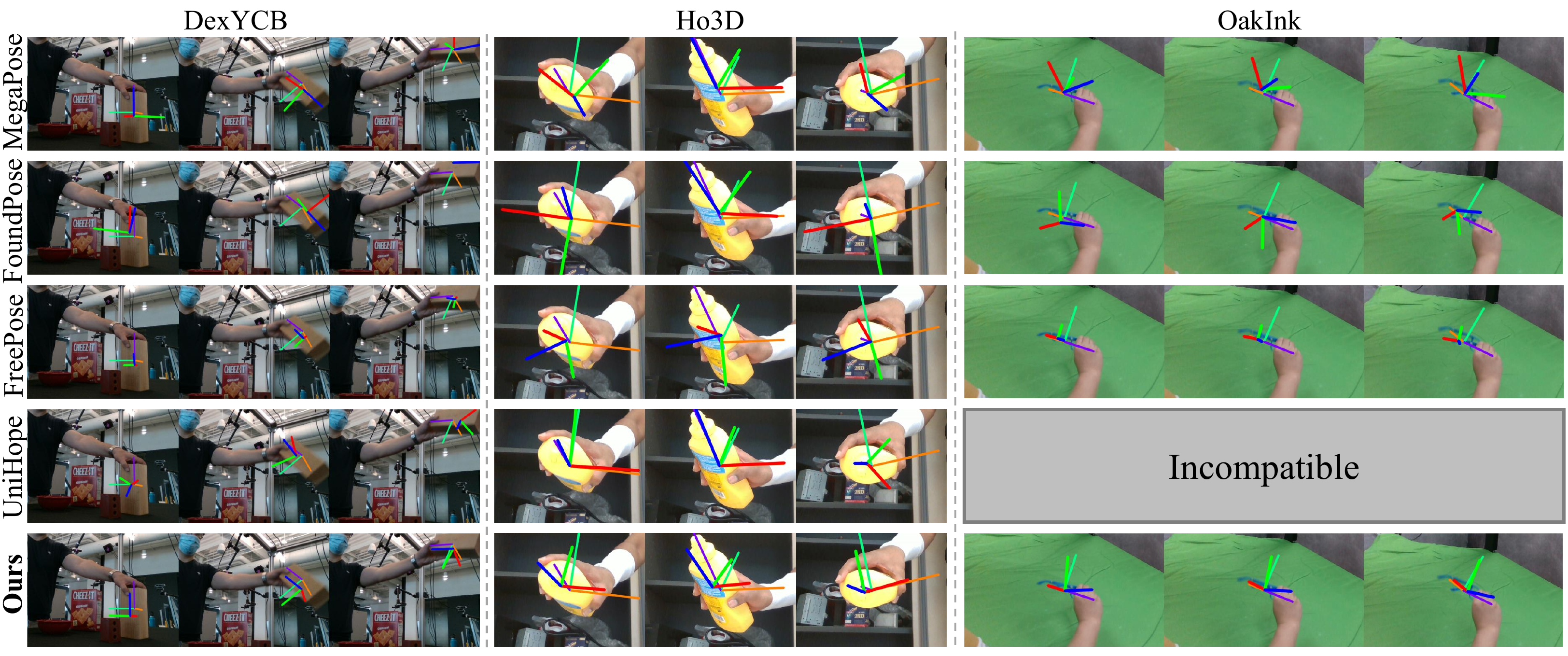}
    \vspace{-6mm}
    \caption{Qualitative results from the state-of-the-art models on three datasets: DexYCB, HO3D-v2, and OakInk-v1. To aid visualization, images are zoomed to highlight the objects.}
    \label{fig:quali}
\end{figure}

Fig.~\ref{fig:quali} and Tab.~\ref{tab:main} present quantitative and qualitative comparisons with existing approaches.
Overall, our method achieves the best or highly competitive performance across most metrics, and consistently obtains the best results for RRE, TCC$_\mathrm{R}$, and TCC$_\mathrm{T}$ on all the datasets, demonstrating strong temporal consistency in relative object motion.
On HO3D, where sequences are substantially longer than in other datasets, absolute metrics such as ARE and ATE are more affected by the accumulated drift when relative predictions are composed over time.
Nevertheless, our method remains competitive, showing that the proposed consistency regularization and anchoring effectively suppress cumulative error.
On OakInk, we further visualize three consecutive frames to highlight the temporal stability.
While the prior methods exhibit noticeable frame-to-frame flickering, our method produces visibly smoother and more stable predictions without any external guidance.

A key advantage of our approach is the inference efficiency.
Unlike render-and-match methods, it does not require mesh rendering or template matching at test time.
UniHOPE also achieves fast inference, but its object prediction depends on regressing the 2D projections of 3D object corner keypoints and solving PnP, which ties it to object meshes and the object taxonomy seen during training, limiting zero-shot generalization to unseen object classes such as those in OakInk.
Template-based methods such as MegaPose~\cite{labbe2022megapose}, FoundPose~\cite{ornek2024foundpose}, and FreePose can further suffer from ambiguity, especially for symmetric, textureless, or heavily occluded objects.
In contrast, our method predicts the relative object motion directly from dense 3D geometry and complementary hand cues, making it more robust to symmetry, texture ambiguity and severe hand occlusion, as reflected in both Fig.~\ref{fig:quali} and Tab.~\ref{tab:main}.

\noindent \textbf{Ablation Study.}
Next, we conduct an ablation study to evaluate the effectiveness of our pipeline design in Tab.~\ref{tab:abl}.
We first compare object-only, hand-only, and full versions of our method to assess the contribution of each component.
Object-only tracking performs reasonably when sufficient object surface is visible, but degrades substantially under severe hand occlusion or limited geometric support.
Hand-only tracking is more robust to object occlusion, but becomes less reliable when the articulated hand motion or re-grasping violates the rigid-motion assumption.
These results confirm that object geometry and hand motion provide complementary information and should be combined adaptively for robust hand-held object tracking.

\begin{wraptable}{r}{0.5\columnwidth}
\vspace{-5mm}
\caption{Ablation on each module of our pipeline.}
\label{tab:abl}
\centering
\vspace{-3mm}
\setlength{\tabcolsep}{2.5pt}
\resizebox{\linewidth}{!}{
\begin{tabular}{l|cccccc}
\toprule
    Variants & ARE & ATE & RRE & RTE & \TCCR & \TCCT  \\ 
    \midrule
    only Object ICP ($\alpha=0$) & 35.02 & 3.08 & 1.97 & 0.6 & 0.37 & 0.57 \\
    only Hand ($\alpha=1$) & 30.65 & - & 2.64 & - & 0.38 & - \\
    Ours (VGGT) & 23.72 & 4.52 & 0.81 & 0.9 & 0.49 & 0.47 \\
    Ours (HaWoR) & 19.88 & 3.08 & 1.52 & 0.6 & 0.56 & 0.57\\
    Ours (w/o anchor) & 24.48 & 3.08 & 1.53 & 0.6 & 0.56 & 0.57\\
    Ours (GT Mask) & 21.27 & 2.44 & 1.57 & 0.43 & 0.55 & 0.62 \\
    \bottomrule
\end{tabular}
}
\vspace{-5mm}
\end{wraptable}

\begin{figure}[t]
    \centering
    \includegraphics[clip, width=\linewidth]{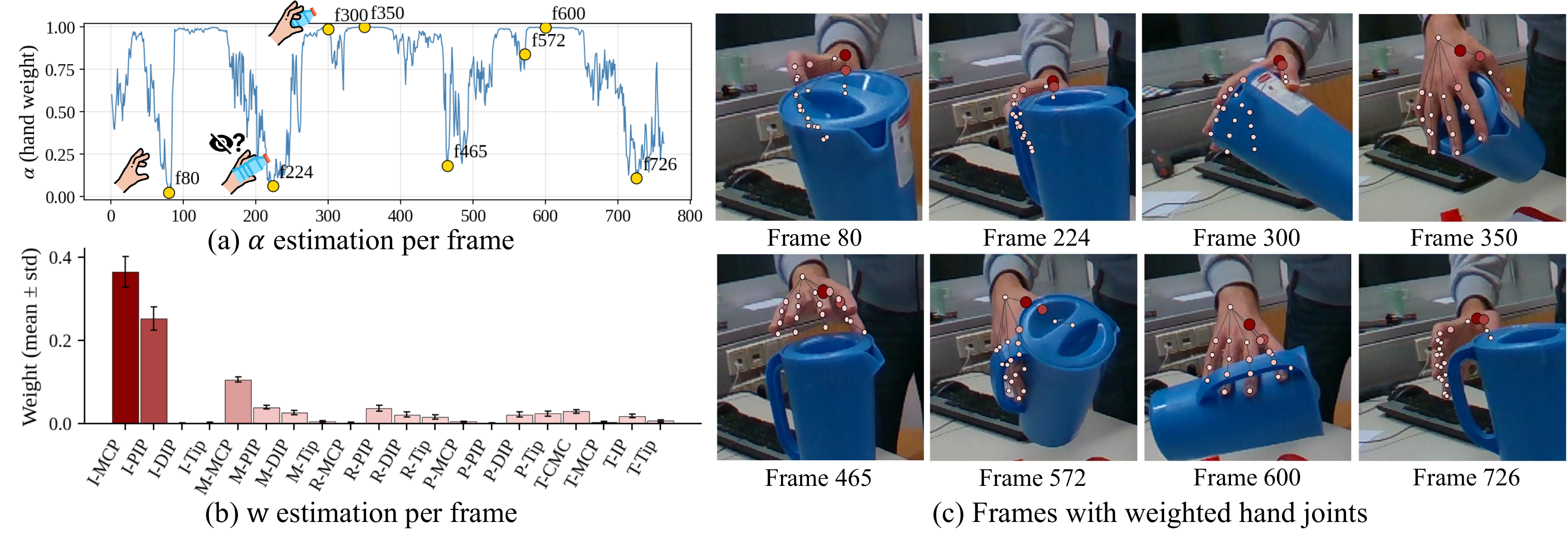}
    \vspace{-6mm}
    \caption{Analysis on the predicted $\alpha$ and $w$ along with the input frames.}
    \label{fig:joint}
\end{figure}

We further analyze the predicted blending coefficient $\alpha$ and joint weights $\mathbf{w}$ in Fig.~\ref{fig:joint}.
As shown in Fig.~\ref{fig:joint}(a), $\alpha$ varies significantly across frames, reflecting the model's adaptive preference between object- and hand-based cues.
When the hand is not yet in stable contact or provides weak evidence for object motion, $\alpha$ remains low and the model relies more on object geometry.
When the hand firmly grasps the object and heavily occludes it, $\alpha$ becomes high, indicating greater reliance on the hand cue.
Fig.~\ref{fig:joint}(b,c) further shows that higher weights are assigned to joints near the thumb and index contact region, while lower weights are imposed to joints dominated by finger articulation.
This supports our weighted Procrustes design, which emphasizes joints that are more reliable for object-aware motion estimation.

Morevoer, we evaluate the effect of annotation quality by replacing the SAM3-based mask with ground-truth annotations.
Using ground-truth mesh and mask improves translation-related metrics, including ATE, RTE, and $\mathrm{TCC}_\mathrm{T}$, indicating that more accurate visible regions and initialization stabilize the translation trajectory.
Furthermore, using VGGT yields slightly lower translation-related metrics, which shows that Flow3R has better potential for capturing dynamic objects from input frames, although still better than the comparison methods. We can also see HaWoR consistently yields good results comparable to WiLoR. 
In total, these ablations support our design choice of combining object- and hand-derived cues within an adaptive tracking framework, and robustness to the choice of the models for object and hand.

\begin{figure}[t]
    \centering
    \includegraphics[clip, width=\linewidth]{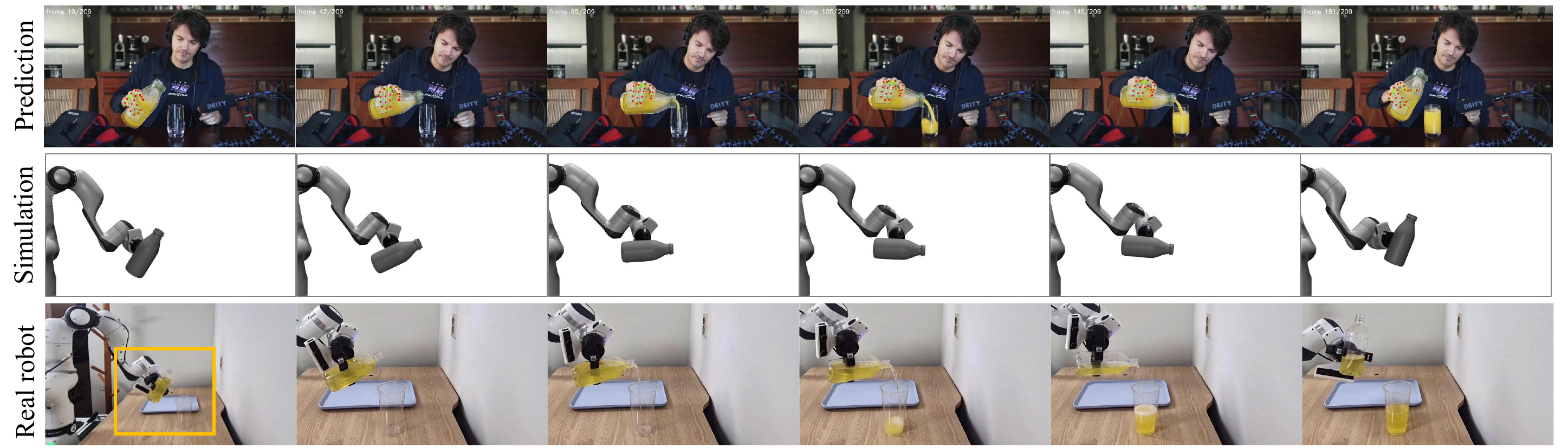}
    \vspace{-6mm}
    \caption{Robot manipulation from internet videos. We overlay the predicted hand joints and mesh of the input frames, and transfer the trajectory to the robot. Our predicted trajectory can be directly utilized without any robot-specific trajectory optimization.}
    \label{fig:robot}
\end{figure}

\subsection{Application: Robot Manipulation}

We directly apply our 6D pose estimates to robotic manipulation. Mapping visual demonstrations to physical robot actions remains challenging due to the need for precise object tracking. 
Our feed-forward architecture generate temporally consistent and accurate pose sequences, eliminating the need for additional robot-specific post-processing~\cite{ponimatkin20256d}. By streaming predicted poses directly to the controller, our method reduces computational overhead and simplifies real-world deployment.

\textbf{Direct Trajectory Retargeting.} Let $T_{\text{cam}}^{\text{obj}}(t) \in SE(3)$ denote the predicted object pose at time $t$. With a known camera-to-base transformation $T_{\text{base}}^{\text{cam}}$ and a fixed grasp offset $T_{\text{eef}}^{\text{obj}}$, the target end-effector pose $T_{\text{base}}^{\text{eef}}(t)$ is simply computed as below:
\begin{equation}
    T_{\text{base}}^{\text{eef}}(t) = T_{\text{base}}^{\text{cam}} \cdot T_{\text{cam}}^{\text{obj}}(t) \cdot (T_{\text{eef}}^{\text{obj}})^{-1}.
\end{equation}
This sequence is directly fed into an Operational Space Controller (OSC) driving a Franka Panda robot, avoiding complex optimal control formulations.

\textbf{Execution in the Real World.}
We first validate the kinematic stability of these raw trajectories in the \texttt{robosuite}~\cite{zhu2020robosuite} simulation environment before directly deploying them on the physical Franka Panda hardware. We validate this streamlined pipeline by imitating unconstrained \textit{Internet videos} and replaying the motion in the real-world. Figs.~\ref{fig:teaser} and \ref{fig:robot} show that the physical robot successfully reproduces fine-grained dynamic motions by directly using the target poses predicted by ComPose. Since we already have camera pose from Flow3R, our method can also directly be applied to egocentric video. More results and further details are provided in supplementary material section and video results.

\section{Conclusion}
\label{sec:conclusion}
We present a novel hand-held object tracking framework that combines object geometry and hand motion under hand-object interaction.
By adaptively fusing object- and hand-based cues, our method remains effective in challenging cases where either source alone is insufficient.
Extensive experiments across multiple benchmarks demonstrate the effectiveness of the proposed formulation.
Moreover, our temporally smooth trajectories can be directly transferred to downstream real-world robot manipulation, enabling stable execution without complex post-hoc optimization.

\noindent \textbf{Limitations.}
Our method is most challenged when both object geometry and hand cues become unreliable, such as under severe occlusion, motion blur, or degraded image quality.
Although we handle various grasp type on our training, it can struggle under unseen grasp distributions, where the model may fail to identify the most informative joints for hand-based motion estimation.
In addition, our anchoring strategy reduces cumulative rotation drift, but its residual drift can still remain over long sequences.
One of future works includes explicit modeling of grasp types and hand-object contacts, scaling to richer HOI datasets and more robust long-horizon drift correction.

\clearpage
{
    \small
    \bibliography{neurips_2026}
    \bibliographystyle{splncs04}
}

\clearpage

\appendix

\section{Appendix}
The appendix includes sections as follows: \\
• Section~\ref{sec:more_impl}: More implementation details of our pipeline. \\
• Section~\ref{sec:robot_mani}: Robot manipulation details.\\
• Section~\ref{sec:additional_result}: Additional results.

\subsection{More implementation details of our pipeline}
\label{sec:more_impl}

\noindent \textbf{Network Details.}
\begin{table}[h]
\setlength{\tabcolsep}{2.5pt}
\centering
\caption{\textbf{Detailed network structure of the complementary pose estimation module.}
The input consists of DINOv2 patch features, object masks, 2D/3D hand joints over a temporal window of $N$ frames. Input images are center-cropped and resized to $224\times224$ ($H, W=224$).
Here, $B$ denotes the batch size, $N$ the number of sampled frames, $T=H'W'$ the number of patch tokens, and $D=384$ the token dimension. 
The module outputs pair-wise joint weights $\mathbf{w}$, blending coefficients $\alpha$, and per-frame translation residuals $\boldsymbol{\delta}$.}
\label{tab:network_structure_tracking}
\resizebox{\linewidth}{!}{
\begin{tabular}{c|c|c|c}
\toprule
\textbf{IDX} & \textbf{STRUCTURE} & \textbf{INPUT} & \textbf{OUTPUT} \\
\toprule
\multicolumn{4}{c}{\textbf{Inputs and Preprocessing}} \\
\toprule
1-1 & DINOv2 feature extraction & Image $(B \times N \times 3 \times H \times W)$ & $B \times N \times H' \times W' \times 384$ \\
1-2 & Mask fusion & DINO feat. + mask $(B \times N \times H' \times W' \times 384)$ & $B \times N \times T \times 384$ \\
1-3 & Frame positional encoding & OUT1-2 & $B \times N \times T \times 384$ \\
1-4 & Joint token initialization & Joint2D $(B \times N \times 21 \times 2)$, Joint3D $(B \times N \times 21 \times 3)$ & $B \times N \times 21 \times 384$ \\
\toprule
\multicolumn{4}{c}{\textbf{Image Encoder}} \\
\toprule
2-1 & FrameAttention & OUT1-3 & $B \times N \times T \times 384$ \\
2-2 & GlobalAttention & OUT2-1 & $B \times N \times T \times 384$ \\
2-3 & FrameAttention & OUT2-2 & $B \times N \times T \times 384$ \\
2-4 & GlobalAttention & OUT2-3 & $B \times N \times T \times 384$ \\
2-5 & FrameAttention & OUT2-4 & $B \times N \times T \times 384$ \\
2-6 & GlobalAttention & OUT2-5 & $B \times N \times T \times 384$ \\
2-7 & FrameAttention & OUT2-6 & $B \times N \times T \times 384$ \\
2-8 & GlobalAttention & OUT2-7 & $B \times N \times T \times 384$ \\
2-9 & Mean pooling over patches & OUT2-8 & $B \times N \times 384$ \\
\toprule
\multicolumn{4}{c}{\textbf{Joint Encoder}} \\
\toprule
3-1 & CrossAttention (joint2D, image) & OUT1-4, OUT2-8 & $B \times N \times 21 \times 384$ \\
3-2 & Joint GlobalAttention & OUT3-1 & $B \times N \times 21 \times 384$ \\
3-3 & CrossAttention (joint2D, image) & OUT3-2, OUT2-8 & $B \times N \times 21 \times 384$ \\
3-4 & Joint GlobalAttention & OUT3-3 & $B \times N \times 21 \times 384$ \\
\toprule
\multicolumn{4}{c}{\textbf{Prediction Heads}} \\
\toprule
4-1 & Rotation weight head & Pair-averaged joint tokens & $B \times 21$ \\
4-2 & Softmax($\tau$) & OUT4-1 & Pair-wise weight $\mathbf{w}_{ij} \in \mathbb{R}^{21}$ \\
4-3 & Alpha head & Concatenated frame pools $(i,j)$ & $B \times 1$ \\
4-4 & Sigmoid & OUT4-3 & Pair-wise $\alpha_{ij} \in [0,1]$ \\
4-5 & Centroid encoder & Centroid $\mathbf{c}$ $(B \times N \times 3)$ & $B \times N \times 96$ \\
4-6 & Translation head & Frame pool + centroid encoding & $\boldsymbol{\delta} \in B \times N \times 3$ \\
\toprule
\multicolumn{4}{c}{\textbf{Geometric Outputs}} \\
\toprule
5-1 & Weighted Procrustes & $J_i$, $J_j$, $\mathbf{w}_{ij}$ & $R_{\mathrm{hand},ij} \in \mathbb{R}^{3 \times 3}$ \\
5-2 & SLERP fusion & $R_{\mathrm{obj},ij}$, $R_{\mathrm{hand},ij}$, $\alpha_{ij}$ & $\hat{R}_{ij} \in \mathbb{R}^{3 \times 3}$ \\
5-3 & Residual translation & $\mathbf{c}_i + \boldsymbol{\delta}$ & $\hat{\mathbf{t}}_i \in \mathbb{R}^{3}$ \\
\bottomrule
\end{tabular}
}
\end{table}

\begin{figure}[t]
    \centering
    \includegraphics[clip, width=\linewidth]{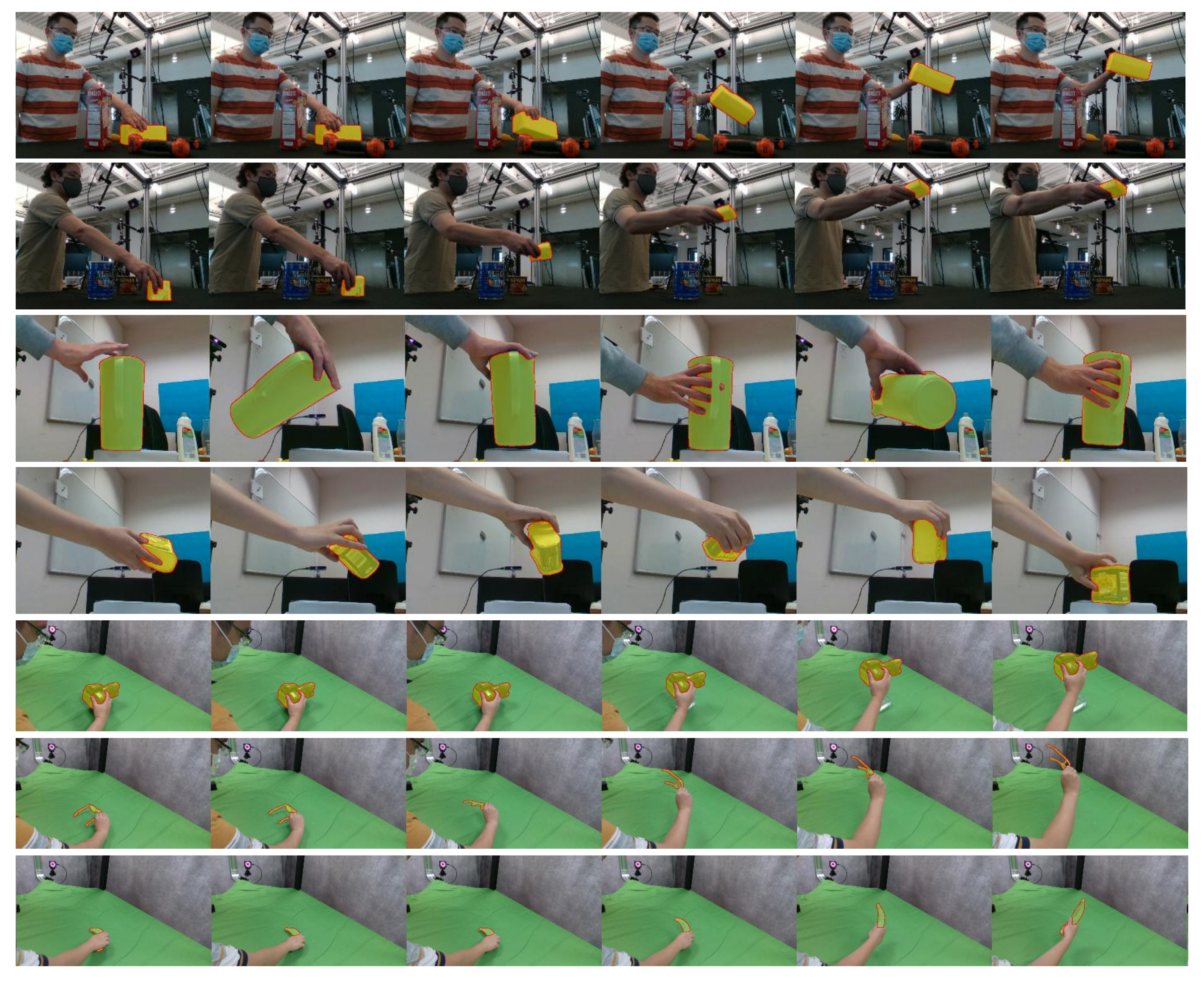}
    \caption{Visualization of mask tracking using SAM3.}
    \label{fig:supple_sam3}
\end{figure}

\noindent \textbf{Object Mask Extraction.} 
To initialize the semantic identity of target objects, we automatically extract per-frame segmentation masks using SAM3~\cite{carion2025sam3segmentconcepts}. We visualize the example results on Fig.~\ref{fig:supple_sam3}. Specifically, we leverage its zero-shot grounding capabilities by providing class-specific text prompts for each video sequence. Among multiple candidate instances, we identify the target object by selecting the mask whose centroid is closest to the hand, effectively isolating the interacting object from the background. These masks are then bidirectionally propagated across the sequence to maintain temporal consistency. To ensure data quality and computational efficiency, masks containing fewer than 100 pixels are discarded as noise, and sequences exceeding 300 frames are subjected to uniform temporal subsampling.

\begin{figure}[t]
    \centering
    \includegraphics[clip, width=\linewidth]{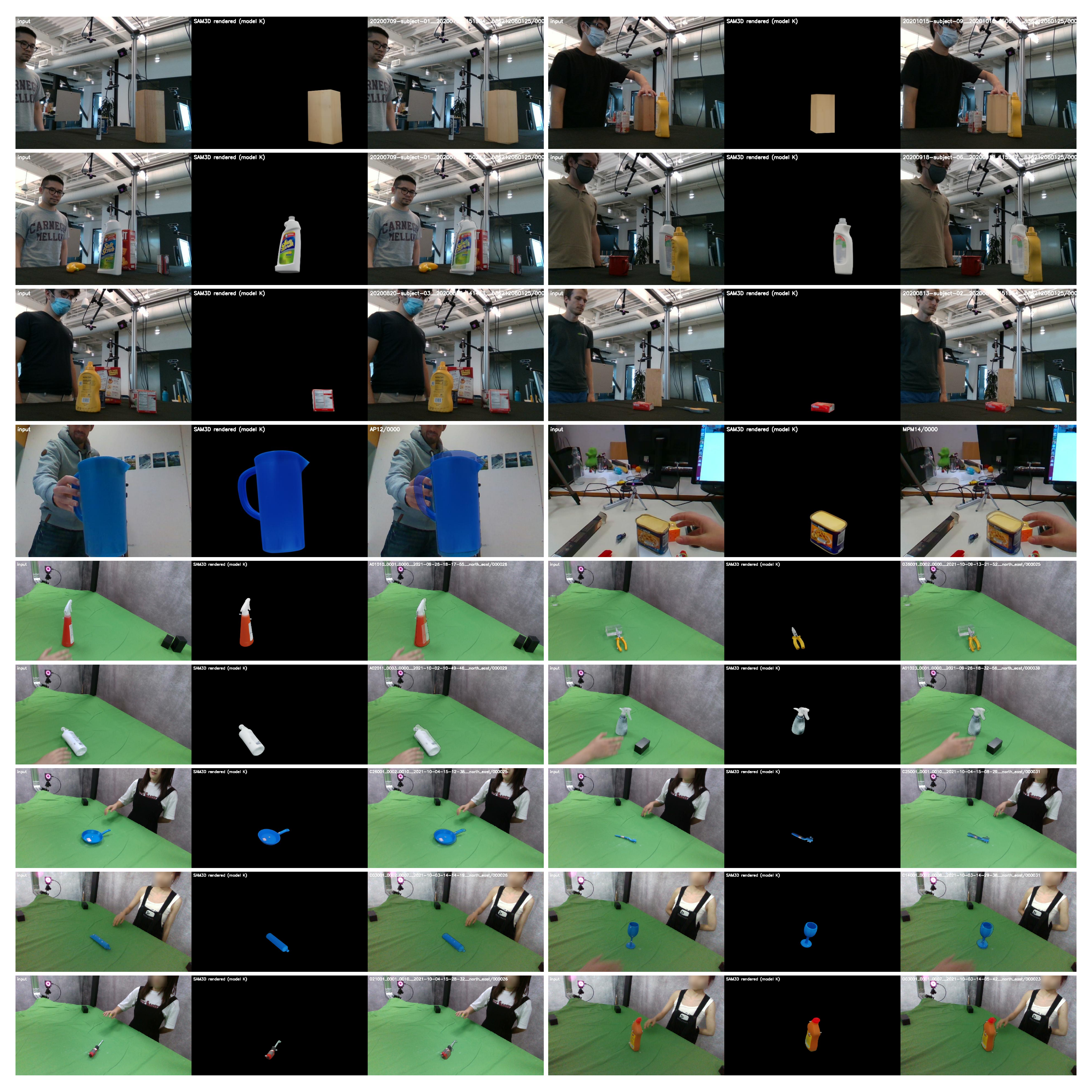}
    \caption{Visualization of mesh reconstruction and pose initialization using SAM3D.}
    \label{fig:supple_sam3d}
\end{figure}

\noindent \textbf{Mesh Reconstruction.}
We reconstruct the initial object mesh using SAM3D-Object~\cite{chen2025sam}, initialized from the SAM3 masks described above.
As shown in Fig.~\ref{fig:supple_sam3d}, the model produces reasonable object meshes across diverse categories, even without object-specific templates.
However, when the reconstructed mesh is projected back to the image, we occasionally observe slight misalignment with the input object.
This behavior is consistent with public user reports on SAM3D-Object that highlight the difficulty of obtaining perfectly aligned renderings from the reconstructed 3D scene, especially when camera alignment is imperfect~\cite{chen2025sam}. 
Because our work does not rely on additional depth-aware refinement, we use the reconstructed mesh directly and leave tighter image-space alignment to future work.

\noindent \textbf{Choice of Models for Object and Hand.}
We use WiLoR~\cite{potamias2025wilor} as the default hand pose estimator because our method mainly relies on accurate frame-level hand pose cues for pairwise object motion estimation.
HaWoR~\cite{zhang2025hawor} is designed for world-space hand motion reconstruction from egocentric videos and incorporates additional temporal components such as camera trajectory estimation, making it less directly aligned with our setting.
As a result, WiLoR provides a simpler and more efficient choice in practice, while our ablation study shows that the overall formulation remains compatible with HaWoR.

For object geometry, we use Flow3R~\cite{cong2026flow3r} as the default 3D foundation model.
Flow3R improves the performance of feed-forward visual geometry prediction with optical-flow-based supervision and is particularly effective on dynamic scenes and moving objects.
This property is especially relevant to our setting, since manipulated objects in hand-object interaction are inherently dynamic.
At the same time, our framework is not tied to a specific 3D foundation model.
As shown in the ablation study Tab.~\ref{tab:abl}, alternative backbones such as VGGT~\cite{wang2025vggt} can also be used, while Flow3R provides slightly stronger performance in our experiments.

After merging those cues with ComPose, we further apply a simple temporal smoothing to our predicted trajectories. Specifically, we use sliding-window quaternion averaging for rotations and moving-average filtering for translations. 

\noindent \textbf{Dataset Details.}

\textit{Training datasets.}
We use the DexYCB~\cite{chao2021dexycb} S3 training and validation split for supervised training.
This split contains 17 objects across 850 sequences.
Since each interaction sequence is captured from 8 synchronized camera views, DexYCB provides diverse observations of the same hand-object motion under varying viewpoints.
We additionally use HOI4D~\cite{liu2022hoi4d}, which contains approximately 2,000 egocentric hand-object interaction sequences and complements DexYCB with more diverse manipulation dynamics and visual appearances.

\textit{Test datasets.}
For in-domain evaluation, we use the DexYCB S3 test split, which contains 3 unseen objects and 150 sequences.
This split evaluates generalization to previously unseen grasped objects under the same capture setup and annotation protocol.
For cross-dataset evaluation, we use the HO-3D v2 evaluation split~\cite{hampali2020honnotate}, which contains 13 sequences across 10 YCB objects.
We further evaluate on the official test split of OakInk-v1~\cite{yang2022oakink}, which contains 57 sequences across 27 objects after filtering.
Since OakInk includes single-hand interactions together with bimanual handover sequences and articulated objects, we exclude the latter cases to match our single-hand tracking setting.
This benchmark evaluates generalization to novel object geometries under unconstrained hand manipulation.

\noindent \textbf{Evaluation Metric Details.}
To evaluate the fidelity of frame-to-frame motion, we report Relative Translation Error (RTE) and Relative Rotation Error (RRE).
Let
\(
\Delta \mathbf{t}_i = \mathbf{t}_{i+1}-\mathbf{t}_i,
\;
\Delta \hat{\mathbf{t}}_i = \hat{\mathbf{t}}_{i+1}-\hat{\mathbf{t}}_i
\)
and
\(
\Delta R_i = R_i^\top R_{i+1},
\;
\Delta \hat{R}_i = \hat{R}_i^\top \hat{R}_{i+1}
\).
We then define
\begin{equation}
\mathrm{RTE} = \frac{1}{N-1}\sum_{i=1}^{N-1}\|\Delta \mathbf{t}_i-\Delta \hat{\mathbf{t}}_i\|_2, \qquad
\mathrm{RRE} = \frac{1}{N-1}\sum_{i=1}^{N-1}\arccos\left(\frac{\mathrm{tr}(\Delta R_i^\top \Delta \hat{R}_i)-1}{2}\right).
\end{equation}
These metrics measure how accurately the predicted trajectory captures translation and rotation changes between consecutive frames, and are therefore sensitive to temporal drift and motion inconsistency.

For absolute pose accuracy, we additionally report Absolute Trajectory Error (ATE) for translation and Absolute Rotation Error (ARE):
\begin{equation}
\mathrm{ATE} = \sqrt{\frac{1}{N}\sum_{i=1}^{N}\|\mathbf{t}_i-\hat{\mathbf{t}}_i\|_2^2}, \qquad
\mathrm{ARE} = \frac{1}{N}\sum_{i=1}^{N}\arccos\left(\frac{\mathrm{tr}(R_i^\top \hat{R}_i)-1}{2}\right).
\end{equation}
While ATE and ARE reflect frame-wise absolute pose accuracy, RTE and RRE provide a complementary view of trajectory quality by directly evaluating relative motion across frames.

\subsection{Robot Manipulation Details}
\label{sec:robot_mani}

\textbf{Initial Object Pose and Grasp Offset.} 
To execute the retargeted trajectory, the rigid transformation between the robot's end-effector and the object, denoted as the grasp offset $T_{\text{eef}}^{\text{obj}}$, must be defined. Because the space of valid grasp poses in $SE(3)$ is theoretically infinite, selecting an appropriate $T_{\text{eef}}^{\text{obj}}$ is important to ensure the entire trajectory remains within the robot's kinematically feasible workspace. An improper offset can cause the manipulator to reach joint limits or kinematic singularities mid-trajectory. 

In our experiments, we manually specified a fixed $T_{\text{eef}}^{\text{obj}}$ for each demonstration. Because the Franka Emika Panda is a kinematically redundant 7-DoF manipulator, it possesses sufficient null-space flexibility to comfortably execute the 6D target trajectories without requiring strict grasp optimization. However, to scale this to a fully automated pipeline, this manual step could be readily replaced. For instance, an off-the-shelf grasp synthesizer (e.g., GraspNet~\cite{mousavian2019graspnet}) could sample candidate poses, allowing the system to automatically select a grasp that maximizes the robot's manipulability index across the entire sequence.

\textbf{Direct Trajectory Execution via OSC.} 
Once the grasp offset $T_{\text{eef}}^{\text{obj}}$ and the camera alignment $T_{\text{base}}^{\text{cam}}$ are anchored, the target end-effector trajectory $T_{\text{base}}^{\text{eef}}(t)$ is fully determined. Unlike prior methods that require computationally expensive trajectory optimization libraries (e.g., \texttt{Aligator} or \texttt{PROXDDP}) to regularize joint velocities and torques from noisy predictions, our pipeline leverages the inherent temporal smoothness and accuracy of our feed-forward model. 
The computed $T_{\text{base}}^{\text{eef}}(t)$ is directly streamed to the Franka Panda's Operational Space Controller (OSC) at the video frame rate (e.g., 30 Hz). The OSC computes the necessary joint torques at 1000 Hz to track each commanded pose, achieving stable physical execution without any offline trajectory-level smoothing.


\subsection{Additional results}
\label{sec:additional_result}
\begin{figure}[t]
    \centering
    \includegraphics[clip, width=\linewidth]{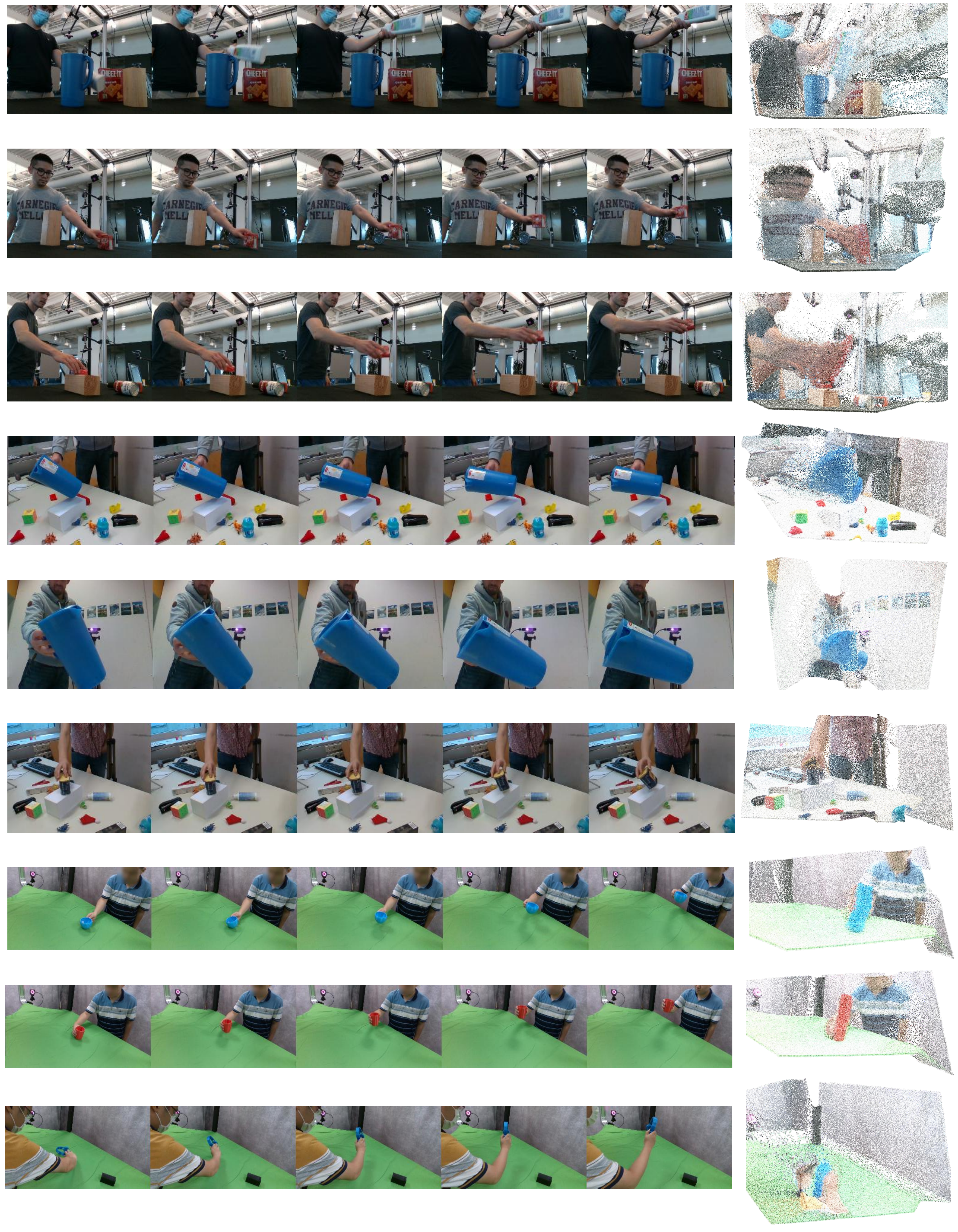}
    \caption{Visualization of point map using Flow3R.}
    \label{fig:supple_flow3r}
\end{figure}

\begin{figure}[t]
    \centering
    \includegraphics[clip, width=\linewidth]{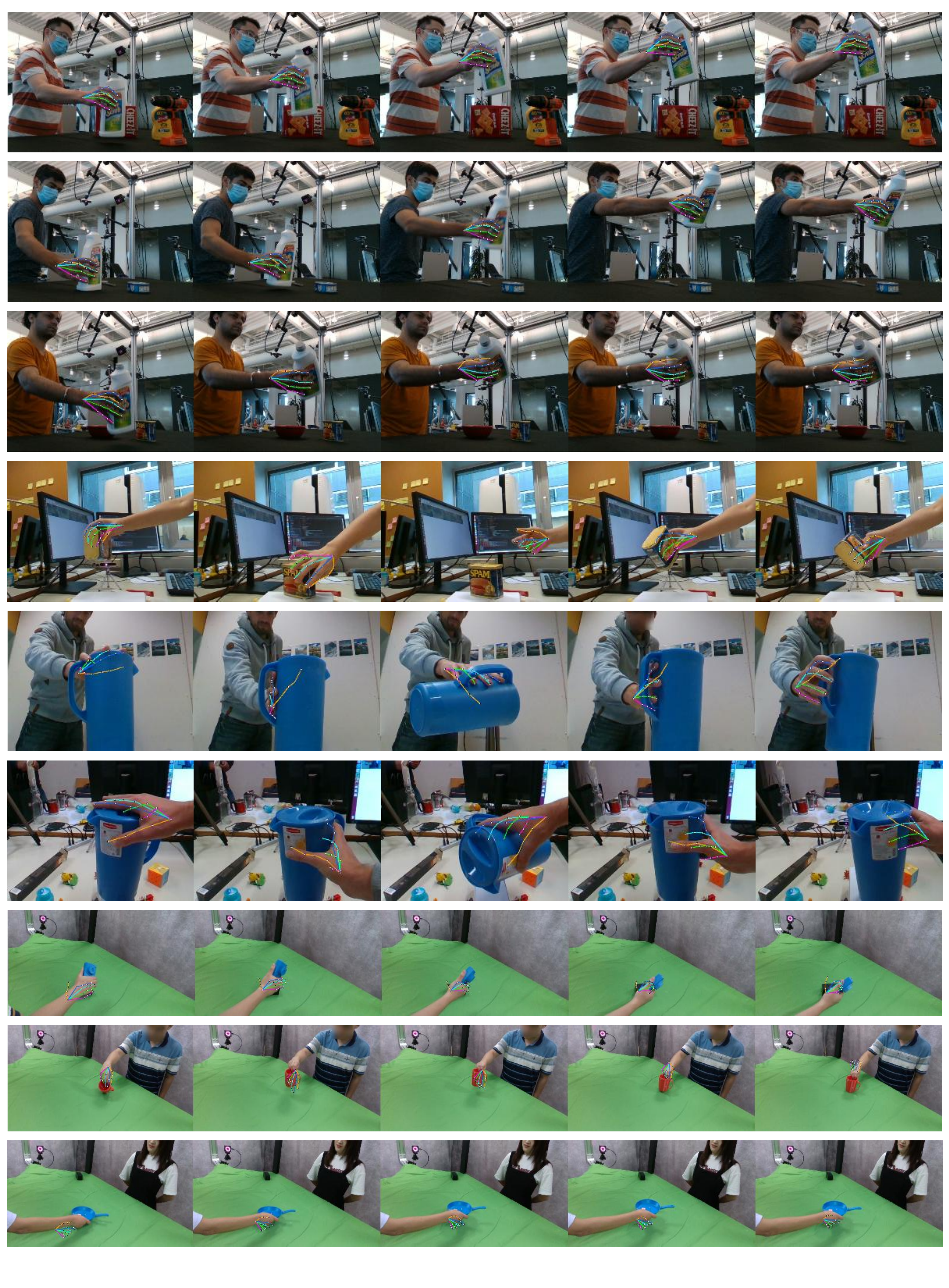}
    \caption{Visualization of 2D hand joints using WiLoR.}
    \label{fig:supple_wilor}
\end{figure}

\begin{figure}[t]
    \centering
    \includegraphics[clip, width=\linewidth]{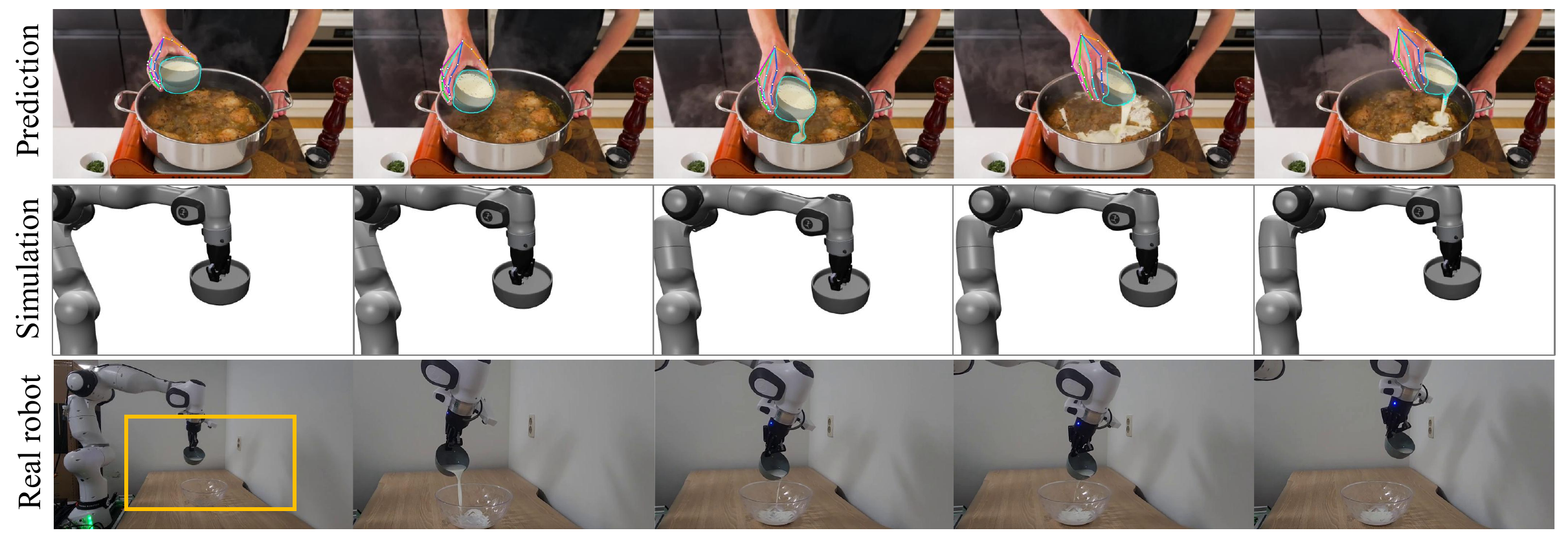}
    \caption{Robot manipulation from Internet video. We overlay the predicted hand joints and mesh on the input frames and transfer the estimated object trajectory to the robot. Our predicted smooth trajectory can be directly used without post hoc trajectory optimization.}
    \label{fig:supple_bowl6}
\end{figure}

We provide results of Flow3R and WiLoR on our test datasets in Fig.~\ref{fig:supple_flow3r}, and~\ref{fig:supple_wilor}. We visualize additional demonstration by transferring trajectory from internet video to robot manipulation. To better match the intended manipulation setup, we conduct the experiment using a small bowl filled with liquid.
As shown by the pouring results(Fig.~\ref{fig:robot}, and~\ref{fig:supple_bowl6}, the transferred trajectory is sufficiently stable to handle not only rigid object motion but also liquid manipulation, which is particularly sensitive to trajectory noise and instability.

Finally, we provide additional qualitative results together with comparisons against prior methods on the benchmark test sets in the attached video.
For fair comparison, we additionally implement the trajectory optimization procedure used in the original FreePose paper~\cite{ponimatkin20256d} and apply it in the FreePose robot simulation setting.
Even without such robot-specific trajectory optimization, our method produces trajectories that transfer effectively to the real robot, further demonstrating the stability and practical usability of the proposed tracking framework.

\clearpage


\end{document}